\documentclass{article}

\usepackage{microtype}
\usepackage{graphicx}
\usepackage{subcaption}
\usepackage{booktabs}
\usepackage{hyperref}
\usepackage{url}
\usepackage{amsmath,amssymb}
\usepackage{amsfonts}
\usepackage{xcolor}
\usepackage{enumitem}
\usepackage{placeins}
\usepackage[accepted]{icml2026}

\icmlshowauthorstrue

\icmltitlerunning{LegalHalluLens}

\begin{document}

\twocolumn[
\icmltitle{LegalHalluLens: Typed Hallucination Auditing and Calibrated\\
Multi-Agent Debate for Trustworthy Legal AI}

\begin{icmlauthorlist}
\icmlauthor{Lalit Yadav}{ind}
\icmlauthor{Akshaj Gurugubelli}{ind2}
\end{icmlauthorlist}
\icmlaffiliation{ind}{Independent Researcher, Sunnyvale, CA, USA}
\icmlaffiliation{ind2}{Independent Researcher, San Diego, CA, USA}
\icmlcorrespondingauthor{Lalit Yadav}{lalitdv94@gmail.com}
\icmlkeywords{legal AI, hallucination evaluation, LLM benchmarking,
compliance risk, AI auditing, trustworthy AI}
\vskip 0.3in
]

\printAffiliationsAndNotice{}

\begin{abstract}
AI systems deployed in legal workflows hallucinate at rates
that aggregate metrics report at $\sim$52\%, but this average
conceals where errors concentrate and in which direction they run,
leaving compliance officers without an actionable signal for
trustworthy deployment. We present \textbf{LegalHalluLens}, an auditing framework with three
components: \textbf{typed hallucination profiles} across four
legally-motivated claim categories (numeric, temporal,
obligation/entitlement, factual) over CUAD~\cite{hendrycks2021cuad};
a \textbf{Risk Direction Index (RDI)} that reduces
omission-versus-invention bias to a single deployment-comparable
scalar; and a \textbf{typed debate pipeline} calibrated to both
magnitudes and directions. Across 510 contracts and 249{,}252
clause-level instances we measure a within-model gap of approximately
38--40~pp between obligation/numeric and temporal claims that aggregate
reporting hides, and show that two systems with matched 52\% rates
can carry opposite RDIs. The debate pipeline reduces fabricated
detections by 45\% with per-category gains tracking the diagnosis,
matching commercial APIs with a substantially smaller backbone (4B active parameters). Typed profiles and RDI surface failure modes that aggregate metrics hide; we further show these diagnostics serve as calibration inputs for multi-agent debate pipelines, where Skeptic challenges and asymmetric gates targeted at measured failure modes outperform generically-tuned debate. The framework supports direction-aware procurement, accountability, and agent design for legal AI deployed in the wild.
\end{abstract}

\section{Introduction}

Legal AI is being deployed in workflows where practitioners make
consequential decisions on the basis of model output,  contract
review, compliance monitoring, regulatory reporting, due diligence,  and where model selection is itself a decision with real legal
exposure.

\paragraph{Why this matters at scale.}
Legal AI errors are asymmetric in who bears the cost. A liability
cap invented by a model and missed in review creates a false risk
ceiling that may be relied on for months. A non-compete scope
qualifier silently dropped may produce an unenforceable clause
that counsel never flags. Trustworthy
deployment requires knowing not just that a system hallucinates at
52\%, but \emph{which clauses}, \emph{in which direction}, and
whether a calibrated intervention can shift that profile at
reasonable cost. The framework we develop is an auditing
instrument: typed profiles and the Risk Direction Index are
derivable from any oracle-bounded legal corpus, supporting
procurement evaluation, post-deployment monitoring, and
direction-aware governance of legal AI. Aggregate hallucination
rates,  the standard reporting practice today,  cannot serve this
role: averaging across claim types conceals exactly the failure
modes that determine legal exposure.

\paragraph{Where prior work stops.}
Prior typological work~\cite{dahl2024fictions, hou2024gaps,
magesh2024hallucinationfree} establishes that legal hallucinations
are not uniform but does not address the contract extraction
setting or collapse directional character into a deployment-comparable
scalar. \S\ref{sec:related} positions this work against each
cluster in detail.

\paragraph{Research questions.}
This paper addresses three questions.

\textbf{RQ1: typed failure ordering.} Do LLMs exhibit
systematically different hallucination rates across legal claim
types, and is this pattern consistent enough across architectures
to function as a reliable evaluation signal? If numeric and
obligation claims fail substantially more than temporal claims
across all tested systems, then any evaluation that averages across
types is concealing the failure rate on the clauses of greatest
legal consequence.

\textbf{RQ2: error direction.} Can the directional character of
content errors, whether a model suppresses obligations present in
the source or asserts ones that are not, be captured in a single
deployment-actionable metric, and does this signal differentiate
systems that aggregate rates cannot?

\textbf{RQ3: typed mitigation.} Does a debate pipeline calibrated
to both the failure magnitudes from RQ1 and the error directions
from RQ2 produce gains concentrated on the highest-failure
categories, and does this calibrated approach enable a small open
model to match or exceed the performance of commercial APIs at
substantially lower inference cost?

\paragraph{Experimental scope.}
We ground the study in structured legal clause extraction using
CUAD v1.0~\cite{hendrycks2021cuad} as an oracle-bounded evaluation
corpus: 510 commercial contracts with 41 expert-annotated clause
types, providing a complete ground-truth oracle in which every
model output is verifiable against the contract text without
external knowledge. We evaluate under full-document context,
measuring the performance ceiling for retrieval-augmented variants
where retrieval errors compound on top of the content failures we
report. \emph{Experiment~1} evaluates four models, two commercial
APIs, one 32B open model, one 70B open model, across all 510
contracts and three runs, yielding 249{,}252 clause-level instances.
\emph{Experiment~2} applies a typed debate pipeline to gemma-4-26B-A4B
(Mixture-of-Experts, 4B active parameters) on a 120-contract matched subset, testing
whether the typed failure profile from Experiment~1 supports a
calibrated and cost-efficient mitigation.

\paragraph{Contributions.}
\begin{enumerate}[leftmargin=*,itemsep=2pt]
\item \textbf{Typed hallucination profiles} (\S\ref{sec:results-exp1}):
a consistent failure ordering
\{numeric, obligation\}~$\gg$~factual~$\geq$~temporal across four
architecturally diverse models, spanning approximately 38--41~pp per model and
not observable under aggregate reporting.
\item \textbf{Risk Direction Index} (\S\ref{sec:direction}): a
signed scalar metric that decomposes content errors into omission
versus invention across typed claim categories, encoding net
directional bias as a single deployment-actionable signal.
\item \textbf{Calibrated multi-agent debate as mitigation}
(\S\ref{sec:results-exp2}): a six-role debate pipeline (Skeptic,
Supporter, Re-extractor, Arbiter, Verifier, Judge) operating on a
baseline extraction, whose Skeptic challenges and Add/Delete gate
asymmetries are derived from the diagnosis above rather than chosen generically. Reduces fabricated detections by 45\% on the matched subset and enables a
4B-active open model to match commercial APIs on composite score
(rank~1 under 4 of 5 weighting schemes) at substantially lower
inference cost.
\end{enumerate}

\section{Related Work}
\label{sec:related}

\paragraph{Legal hallucinations and benchmarks.}
\citet{dahl2024fictions} develop a typology of legal
hallucinations across federal-judiciary tasks (rates between
58\% and 88\% depending on model), arguing that ``not all modes
of hallucination are equally concerning for legal professionals.'' \citet{hou2024gaps} construct
a fine-grained taxonomy of gap categories for machine-generated
legal analysis. \citet{magesh2024hallucinationfree} show that RAG in commercial
legal AI tools does not eliminate hallucinations. We take these as
starting premises; neither addresses contract extraction, and
neither collapses directional character into a deployment-comparable
scalar, the gap our four-category taxonomy
(\S\ref{sec:background}) and Risk Direction Index
(\S\ref{sec:method-rdi}) fill. Legal benchmarks~\cite{guha2023legalbench,
blairstanek2024blt, liu2025contracteval} measure task accuracy
without per-claim-type hallucination stratification;
CUAD~\cite{hendrycks2021cuad} provides expert annotations for
classification, which we repurpose as a hallucination oracle.
Other diagnostics are orthogonal~\cite{enguehard2025lemaj,
demir2025validate, purushothama2025bench}.

\paragraph{General hallucination benchmarks and debate-based mitigation.}
FActScore~\cite{min2023factscore},
HaluBench~\cite{ravi2024halubench},
HalluLens~\cite{bang2025hallulens}, and
PHANTOM~\cite{ji2025phantom} measure factual precision without
claim-type stratification. Multi-agent debate has been studied as
a factuality mechanism~\cite{du2024multiagent, fang2025counterfactual,
liu2025selfdebating, hu2025debategraph} with theoretical motivation
from inference-time scaling~\cite{snell2024scaling, wu2024inference}.
Our contribution is the calibration of Skeptic challenges and
asymmetric gates against the per-category and per-direction failure
modes from Experiment~1; \citet{huang2024cannot} contextualises the
content-correction limit we observe.

\paragraph{Agent design in high-stakes deployment.}
Recent multi-agent debate work tunes Skeptic prompts, gate
thresholds, and aggregation rules generically across all error
types~\cite{du2024multiagent, hu2025debategraph}. We argue this
generic tuning is the wrong default for high-stakes wild
deployment: the appropriate Skeptic challenge depends on which
failure modes the underlying model actually exhibits, and the
appropriate gate asymmetry depends on the directional risk
profile. The typed profiles (\S\ref{sec:method-profiles}) and
RDI (\S\ref{sec:method-rdi}) are designed as calibration inputs
for agent design rather than as standalone benchmark numbers.
Our debate pipeline (\S\ref{sec:method-debate}) instantiates the
recipe: Skeptic challenges are derived from the per-type failure
profile, and Add/Delete gate asymmetry is set by the measured
FAR-vs-FRR profile. To our knowledge this is the first multi-agent
extraction pipeline whose components are calibrated from measured
per-failure-mode diagnostics rather than chosen generically.

\section{Background}
\label{sec:background}

We briefly introduce the domain knowledge needed to follow our
contributions: the verification structure of legal text that
motivates our four-category claim taxonomy, and the metrics and
notation we use throughout.

\subsection{Verification Structure of Contract Text}

Commercial contracts contain claims of fundamentally different
verification character. A claim of the form ``the cap on
liability is \$5{,}000{,}000'' has a single numeric value whose
correctness is decidable by direct comparison against the source.
A claim of the form ``the agreement terminates on December 31, 2024''
similarly reduces to verbatim string comparison. By contrast, a
claim such as ``the supplier shall, except as provided in \textit{Section
4.2}, indemnify the buyer against third-party claims arising from
products manufactured before the effective date'' carries multiple
semantic elements that must all be preserved: the modal verb
(\emph{shall}), the carve-out (\emph{except as provided}), the
scope (\emph{products manufactured before}), and the temporal
anchor. Identity claims such as governing law or counterparty name
are short and structurally simple but rely on the model resisting
its parametric prior of common law jurisdictions.

These four verification regimes correspond to the categories we
use throughout the paper: \textbf{numeric}, \textbf{temporal},
\textbf{obligation/entitlement}, and \textbf{factual}. The
categories are defined by primary verification challenge rather
than by document type, so the same categorisation transfers to
any legal extraction task in which model claims can be checked
against a source.

\subsection{Metrics and Notation}

Let $D$ denote a legal document, $c_i$ a claim type from a fixed
inventory $\mathcal{C}$, and $M$ an extraction model. For each
$(D, c_i)$ pair the model outputs either a clause extraction or a
``not present'' decision. Per-instance outcomes form a confusion
matrix $\{\mathrm{TP},\mathrm{FP},\mathrm{FN},\mathrm{TN}\}$
relative to the CUAD oracle (TP $=$ correctly detected as present;
FP $=$ fabricated, asserted present when absent; FN $=$ missed,
present but called absent; TN $=$ correctly absent). A judge then
labels each TP as \emph{supported} or \emph{contradicted},
together with a categorical
\texttt{mismatch\_type} when an error is identified. We report:

\begin{itemize}[leftmargin=*,itemsep=1pt]
\item $\mathbf{FAR} = \mathrm{FP}/(\mathrm{FP}+\mathrm{TN})$
  \hfill\textit{false-acceptance: invents absent clauses}
\item $\mathbf{FRR} = \mathrm{FN}/(\mathrm{FN}+\mathrm{TP})$
  \hfill\textit{false-rejection: misses present clauses}
\item $\mathbf{Acc} = (\mathrm{TP}+\mathrm{TN})/N$
  \hfill\textit{detection accuracy}
\item $\mathbf{Hal_{TP}} = \mathrm{contradicted}/\mathrm{TP}$
  \hfill\textit{content-quality among detections (Exp.~1)}
\item $\mathbf{Hal_{Gen}} = (\mathrm{contradicted}+\mathrm{FP})/(\mathrm{TP}+\mathrm{FP})$
  \hfill\textit{quality of all generated outputs (Exp.~2)}
\item $\mathbf{JEq} = \mathrm{supported}/(\mathrm{TP}+\mathrm{FN})$
  \hfill\textit{end-to-end correctness}
\item $\mathbf{RDI}$: see \S\ref{sec:method-rdi}
\end{itemize}

The two hallucination metrics differ in scope.
$\mathrm{Hal_{TP}}$ measures content correctness \emph{conditional
on detection}: of clauses the model said it found, what fraction
had wrong content? It isolates the failure mode where the right
clause is located but the extracted text is incorrect, and is the
primary signal for typed profiles in Experiment~1.
$\mathrm{Hal_{Gen}}$ is stricter: of \emph{everything the model
emitted as a clause}, what fraction was wrong, counting both
content contradictions \emph{and} fabrications? Because $\mathrm{Hal_{Gen}}$
penalises FPs, it is the appropriate metric for evaluating a
mitigation pipeline that reduces fabrication, and we use it as the
content-quality column in the matched-subset leaderboard
(Experiment~2). FAR and FRR are detection-level metrics; the two
hallucination metrics measure content quality. All four are
complementary and reported together in their respective benchmarks.

\section{Method}
\label{sec:method}

We describe three components: (i) the typed hallucination profile,
(ii) the Risk Direction Index, and (iii) the typed debate pipeline.
The first two are evaluation procedures; the third is a mitigation
mechanism informed by their output.

\subsection{Typed Hallucination Profiles}
\label{sec:method-profiles}

For a model $M$ evaluated on a corpus $\mathcal{D}$, we partition
all clause-level outputs by claim category $c_i \in \{\mathrm{numeric},
\mathrm{temporal}, \mathrm{obligation}, \mathrm{factual}\}$ and
report $\mathrm{Hal_{TP}}(M, c_i)$ stratified per category. The
within-model typed gap is defined as
\[
\mathrm{Gap}(M) = \max_{c_i} \mathrm{Hal_{TP}}(M, c_i) -
                  \min_{c_i} \mathrm{Hal_{TP}}(M, c_i).
\]
A model with a large $\mathrm{Gap}(M)$ has hallucination rates that
vary substantially across claim categories, which means aggregate
$\mathrm{Hal_{TP}}$ averages claim types whose deployment
consequences differ.

\subsection{Risk Direction Index (RDI)}
\label{sec:method-rdi}

The judge returns a \texttt{mismatch\_type} label from a fixed
inventory: \texttt{none}, \texttt{numeric}, \texttt{temporal},
\texttt{obligation}, \texttt{scope}, \texttt{missing\_condition},
\texttt{extra\_condition}, \texttt{other}. Two of these labels carry
directional meaning: \texttt{missing\_condition} (the model omits a
qualifier present in ground truth) and \texttt{extra\_condition} (the
model asserts a qualifier absent from the source). RDI is defined as
\[
\mathrm{RDI}(M) = \frac{\mathrm{p}_{\mathrm{extra}}(M) -
  \mathrm{p}_{\mathrm{missing}}(M)}{100},
\]
where $\mathrm{p}_{\mathrm{extra}}$ and $\mathrm{p}_{\mathrm{missing}}$
are the percentages of contradicted findings carrying each label.
Positive values indicate invention-heavy failure (overstates);
negative values indicate omission-heavy failure (understates).

The directional concept is recognised qualitatively in prior
work~\cite{dahl2024fictions, hou2024gaps}; what we add is the
operationalisation as a single signed scalar derivable from labels
the judge produces already. RDI is intended as a directional signal
rather than a cardinal measure of risk magnitude: scope errors
account for 62--71\% of contradictions in our data and compress
the directional component. The empirical claim
(\S\ref{sec:results-exp1}) is that RDI cleanly separates two systems
with matched aggregate $\mathrm{Hal_{TP}}$.

\subsection{Typed Debate Pipeline}
\label{sec:method-debate}

The debate pipeline (Figure~\ref{fig:pipeline}) operates on a
baseline clause extraction (Figure~\ref{fig:pipeline}, leftmost
node) and is a state machine with six agent roles: a
\emph{Skeptic} that issues typed challenge
questions targeting the failure modes measured in
\S\ref{sec:method-profiles}; a \emph{Supporter} that defends the
extraction using only verbatim contract quotes; a \emph{Re-extractor}
that re-runs extraction from the source when a structural error is
identified; an \emph{Arbiter} that resolves deadlock when agents disagree after all rounds, applying
a conservative policy that preserves the baseline unless contrary
evidence is strong; a \emph{Verifier} that searches the contract
independently and checks definition fit; and a \emph{Judge} that
reads the full debate transcript, Verifier report, and Arbiter
assessment to make all binding content decisions, subject to
asymmetric structural gates. Routing after each Supporter response:
if the Skeptic flagged a structural error in Round~1, the
\texttt{reextract\_node} fires (once only); if both agents agree,
the clause proceeds to Verifier; if rounds remain and agents disagree,
the debate loops; if rounds are exhausted without consensus, the
Arbiter resolves the deadlock before Verifier and Judge. The pipeline
runs for at most two rounds. Three design choices distinguish this
pipeline from generic multi-agent debate.

\textbf{Typed Skeptic challenges.} For numeric claims, the Skeptic
asks whether the value is verbatim in the source or substituted by
a common prior. For obligation claims, it asks whether the modal
verb is preserved and whether all carve-outs are captured. For
temporal claims, it asks whether the value is stated explicitly or
inferred. For factual claims, it asks whether the information comes
from the document or from external knowledge. Full challenge sets
appear in Appendix~\ref{app:skeptic}.

\textbf{The \texttt{reextract\_node}.} When the Skeptic identifies
that the wrong clause was extracted (rather than imprecise content
within the right clause), the pipeline re-runs extraction from the
source rather than debating an answer that cannot be repaired. This
targets structural extraction errors, which are distinct from the
within-clause scope errors that account for 62--71\% of content
contradictions.

\textbf{Asymmetric structural gates.} The Addition Gate
(absent~$\to$~present) requires both Verifier confirmation and
debate consensus before accepting a new detection. The Deletion
Gate (present~$\to$~absent) is blocked when the Verifier confirms
presence, preventing over-conservative removal of real findings.
The asymmetry encodes the FAR~$>$~FRR risk profile measured in
Experiment~1 for high-error claim types.

\section{Experiments}
\label{sec:experiments}

\subsection{Dataset and Oracle}

We use CUAD v1.0~\cite{hendrycks2021cuad}: 510 commercial contracts
with 41 expert-annotated clause types. CUAD is chosen because it
provides a complete ground-truth oracle against which every model
output is verifiable from the contract text alone, no external
knowledge is used at any stage. We map the 41 clause types to four
categories by primary verification challenge (Appendix~\ref{app:mapping}):

\begin{center}\footnotesize
\begin{tabular}{@{}lrl@{}}
\toprule
Category & $n$ & Primary verification challenge \\
\midrule
Numeric    &  5 & Threshold fabrication, unit mismatch \\
Temporal   &  6 & Implied deadline inference \\
Obligation/Ent. & 27 & Modal drift, condition loss \\
Factual    &  3 & Outside-knowledge injection \\
\bottomrule
\end{tabular}
\end{center}

\noindent The Factual and Numeric categories have small $n$;
results for these categories are reported as supporting evidence,
with our central typed-gap claim resting on the Obligation
($n{=}27$) versus Temporal ($n{=}6$) contrast.

\subsection{Models}

\textbf{Experiment 1 (typed profiles benchmark).} Four models at
temperature~$=$~0: gemini-3-flash and gpt-5.2 (commercial
APIs); qwen3-32b (open, 32.8B parameters); llama-3.3-70b (open,
70B). All extract clauses with identical structured-JSON prompts
(Appendix~\ref{app:extract}).

\textbf{Experiment 2 (typed debate mitigation).} Backbone:
gemma-4-26B-A4B (Mixture-of-Experts; 4B active parameters)\footnote{Released under the Apache~2.0 license.}
This model is held out from Experiment~1 to keep the mitigation
study separate from the benchmark, and is selected as the worst
baseline composite score on the matched subset, any improvement
is therefore attributable to the intervention rather than to a
stronger starting point.

\subsection{External Evaluation Judge}

A single external evaluation judge (gemini-2.5-flash,
temperature~$=$~0) scores each extracted clause against CUAD
ground truth under a strict five-criterion rubric: exact numeric
precision, temporal precision, modality match, polarity match,
and exception/carve-out preservation. The judge returns a
\emph{supported / contradicted} verdict and a
\texttt{mismatch\_type} label. The full judge prompt appears in
Appendix~\ref{app:judge}. The same evaluation judge is used for
both experiments and produces every reported $\mathrm{Hal_{TP}}$,
$\mathrm{Hal_{Gen}}$, and RDI value in the paper. This is
distinct from the in-debate Judge node
(\S\ref{sec:method-debate}, Figure~\ref{fig:pipeline}), which
shares the extraction backbone (gemma-4-26B-A4B in Experiment~2) and
adjudicates the Add/Del gates internally to the pipeline; the
in-debate Judge does not score outputs against ground truth and
does not contribute to the reported metrics.

\subsection{Protocol and Scale}

\textbf{Experiment 1.} Three independent runs per model on all 510 contracts.
Nominal opportunities are $510 \times 41 \times 3 = 62{,}730$ per model.
Actual exported totals are 62{,}580 (gemini-3-flash), 62{,}689 (gpt-5.2),
61{,}536 (qwen3-32b), and 62{,}447 (llama-3.3-70b), yielding
\textbf{249{,}252} clause-level instances total. The 0.2--1.9\% shortfall
is contract-correlated rather than random: only 5 contracts fail across
all three qwen3-32b runs (8.6\% of affected contracts), indicating a set
of consistently challenging inputs rather than stochastic dropout
(App.~\ref{app:robustness}).

\textbf{Experiment 2.} A 120-contract matched subset
(run\_id~$=$~1, nominal 4{,}920 opportunities) for direct
baseline-vs-debate comparison.

\section{Results: Typed Hallucination Profiles (Experiment 1)}
\label{sec:results-exp1}

\subsection{Aggregate Rates Cannot Support Legal Deployment Decisions}

 \begin{table}[h]
\caption{Aggregate metrics on the full 510-contract benchmark
(three runs). $\mathrm{Hal_{TP}}$ measures content errors among
detected clauses; $\mathrm{Hal_{Gen}}$ adds fabrications into the
denominator and reorders the models.}
\label{tab:overall}
\footnotesize\centering\setlength{\tabcolsep}{3pt}
\begin{tabular}{@{}lrrrrrr@{}}
\toprule
Model & FAR & FRR & Acc & $\mathrm{Hal_{TP}}$ & $\mathrm{Hal_{Gen}}$ & JEq \\
\midrule
gemini-3-flash & 19.1 & 4.5  & 85.6 & 50.9 & 65.5 & 46.9 \\
gpt-5.2        & 11.8 & 11.6 & 88.3 & 51.8 & 62.4 & 42.6 \\
qwen3-32b      & 13.4 & 10.8 & 87.5 & 52.1 & 63.6 & 42.7 \\
llama-3.3-70b  &  7.7 & 18.0 & 89.0 & 56.5 & 63.7 & 35.7 \\
\bottomrule
\end{tabular}
\end{table}

Table~\ref{tab:overall} illustrates the evaluation problem. Four
architecturally distinct models, two commercial APIs, a 32B open
model, and a 70B open model, fall within a 6 pp
$\mathrm{Hal_{TP}}$ band (50.9--56.5\%). This range is too narrow
to support deployment decisions. A compliance officer comparing
these systems on aggregate $\mathrm{Hal_{TP}}$ would have no
actionable signal.

\subsection{The Typed Failure Ordering Is Consistent and Large}

\begin{figure}[!htbp]
\centering
\includegraphics[width=\columnwidth]{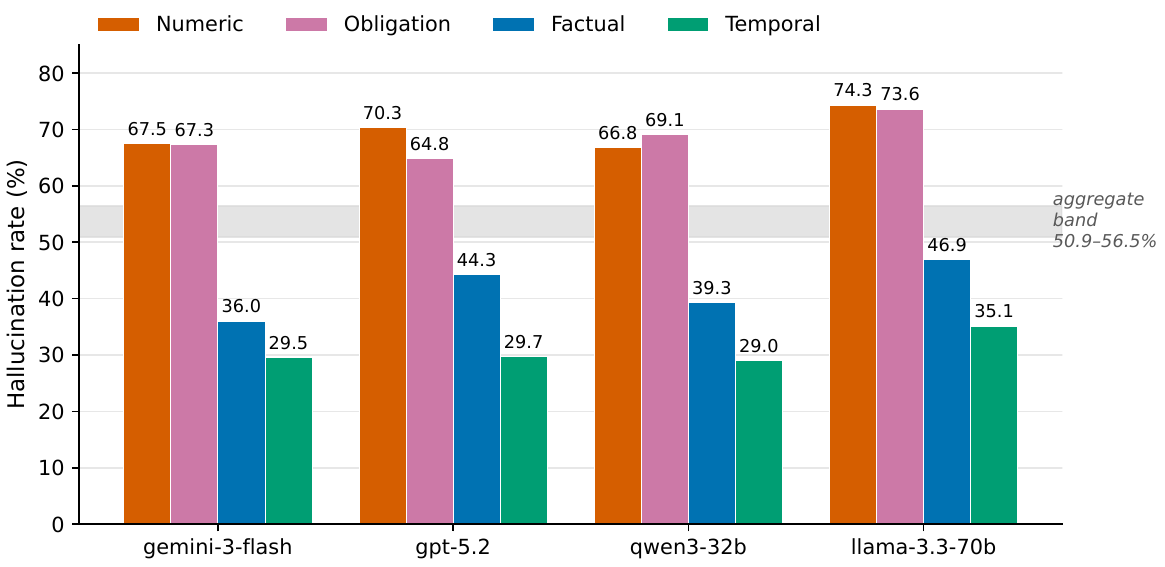}
\caption{Typed hallucination rates on the 510-contract benchmark.
The grey band marks the aggregate $\mathrm{Hal_{TP}}$ cluster (50.9--56.5\%).
Numeric and obligation claims hallucinate at 64.8--74.3\% across
every tested model; temporal claims remain at 29.0--35.1\%. The
resulting within-model gap (approximately 38--41 pp) is not observable under
aggregate reporting.}
\label{fig:typed}
\end{figure}

\begin{table}[h]
\caption{Typed hallucination profiles ($\mathrm{Hal_{TP}}$~\%,
content hallucination among detected clauses). Gap~$=$~max $-$ min across types per model.}
\label{tab:typed}
\scriptsize\centering\setlength{\tabcolsep}{4pt}
\begin{tabular}{@{}lrrrrr@{}}
\toprule
Model & Num. & Obl. & Fact. & Temp. & Gap \\
\midrule
gemini-3-flash & 67.5 & 67.3 & \textbf{36.0} & 29.5 & 38.0 \\
gpt-5.2        & 70.3 & \textbf{64.8} & 44.3 & 29.7 & 40.6 \\
qwen3-32b      & \textbf{66.8} & 69.1 & 39.3 & \textbf{29.0} & 40.1 \\
llama-3.3-70b  & 74.3 & 73.6 & 46.9 & 35.1 & 39.2 \\
\midrule
Range & 66.8--74.3 & 64.8--73.6 & 36.0--46.9 & 29.0--35.1 & 38.0--40.6 \\
\bottomrule
\end{tabular}
\end{table}

Figure~\ref{fig:typed} and Table~\ref{tab:typed} reveal what the
aggregate band conceals. The failure ordering
\{numeric, obligation\}~$\gg$~factual~$\geq$~temporal holds for
every model without exception. A system appearing ``51\%
unreliable'' in aggregate is in fact \textbf{65--74\% unreliable}
on numeric and obligation claims, the categories that determine
liability thresholds, obligation scope, and contract
enforceability, while being only 29--35\% unreliable on temporal
claims.

Two factors explain the disparity. First, obligation clauses
genuinely carry more that can go wrong: modal verbs, trigger
conditions, carve-outs, and scope qualifiers, while a temporal
claim is typically a single verbatim value. Second, our extraction
prompt includes explicit NOTE blocks specifying what does
\emph{not} qualify as each numeric clause type, yet numeric ranks among the two highest-failure types in every
model, never displaced despite explicit prompt guidance:
pretraining priors about common threshold values (``liability caps
are usually \$5M or \$10M'') override explicit instructions, a
finding that bears directly on how far prompt engineering can
compensate for parametric bias.

No single model dominates: qwen3-32b leads on numeric (66.8\%)
and temporal (29.0\%); gpt-5.2 leads on obligation (64.8\%);
gemini-3-flash leads on factual (36.0\%) and end-to-end JEq
(46.9\%). The best choice depends on which claim type is central
to the deployment. Aggregate-based selection can yield the wrong
answer whenever the most consequential claim type for a given
deployment differs from the average. Conservative abstention is
not a safe fallback either: llama-3.3-70b records the lowest FAR
(7.7\%) and highest Acc (89.0\%), but on numeric clauses its FRR
reaches \textbf{52.8\%} and its numeric JEq is only
\textbf{12.1\%}, fewer than 1 in 8 numeric clauses correctly
extracted with correct content. A model silent on liability caps
is not safe for a compliance workflow.

\subsection{The Compliance Direction Problem}
\label{sec:direction}

\begin{figure}[!htbp]
\centering
\includegraphics[width=\columnwidth]{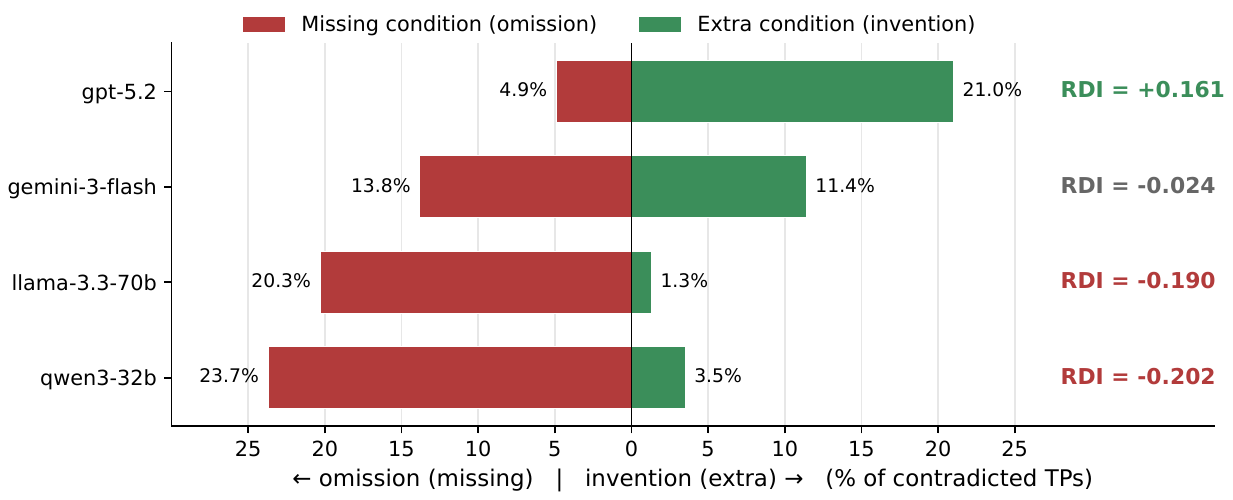}
\caption{Error direction across benchmark models (percentage of
contradicted TP findings). Scope errors dominate universally
(62--71\%), but the residual signal reveals a deployment-critical
distinction: qwen3-32b predominantly omits conditions (23.7\%
missing-condition errors), whereas gpt-5.2 predominantly invents
them (21.0\% extra-condition errors). Both systems report 52\%
aggregate $\mathrm{Hal_{TP}}$. Only the directional decomposition separates their
compliance risk profiles.}
\label{fig:rdi}
\end{figure}

qwen3-32b ($\mathrm{Hal_{TP}} =$ 52.1\%) and gpt-5.2 ($\mathrm{Hal_{TP}} =$ 51.8\%) are
essentially indistinguishable under aggregate evaluation.
Figure~\ref{fig:rdi} shows that they fail in opposite directions.

The underlying distinction is one that compliance practitioners
already reason about and that prior typological
work~\cite{dahl2024fictions, hou2024gaps} has discussed
qualitatively: \emph{do a model's errors tend to suppress
obligations present in the document, or to assert ones that are
not?} These two failure modes have different legal consequences.
A model that drops the ``within 50 miles'' scope qualifier from a
non-compete clause leaves the employer with an unenforceable
overreach that counsel may not flag. A model that invents a
liability cap where none exists creates a false risk ceiling that
materially alters a client's assessment. Both kinds of error score
identically on aggregate $\mathrm{Hal_{TP}}$, but the appropriate remediation and
the exposure carried differ.

The RDI operationalises this distinction using the
\texttt{missing\_condition} and \texttt{extra\_condition} labels
already returned by the judge, it requires no additional
annotation or model calls. The warrant for naming it is not that
the directional concept is novel (it is not, see
\citealp{dahl2024fictions, hou2024gaps}) but that reducing direction
to a single signed scalar lets practitioners compare systems
directly on the question that aggregate $\mathrm{Hal_{TP}}$ cannot answer.

\begin{table}[h]
\caption{RDI and 95\% bootstrap CIs (2{,}000 resamples) for all
four models. gpt-5.2 and qwen3-32b intervals do not overlap,
confirming the directional separation is stable, not run noise.}
\label{tab:rdi}
\centering\footnotesize\setlength{\tabcolsep}{4pt}
\begin{tabular}{@{}lcc@{}}
\toprule
Model & RDI [95\% CI] & Direction \\
\midrule
gpt-5.2        & $+0.161\ [+0.151, +0.170]$ & invents (overstates) \\
gemini-3-flash & $-0.024\ [-0.035, -0.015]$ & near-balanced \\
llama-3.3-70b  & $-0.190\ [-0.198, -0.180]$ & omits (understates) \\
qwen3-32b      & $-0.202\ [-0.212, -0.193]$ & omits most \\
\bottomrule
\end{tabular}
\end{table}

RDI should be read as a directional signal rather than a cardinal
measure of risk. Scope errors (62--71\% of contradictions) compress
the directional variance: many errors are neither clearly omission
nor invention but reflect a wrong semantic aspect. RDI captures only the portion of errors with a clear
directional character. Despite this compression, the signal
cleanly separates qwen3-32b from gpt-5.2, which is the distinction
aggregate $\mathrm{Hal_{TP}}$ cannot make.

In legal workflows where a missed obligation creates liability
(regulatory compliance, covenant monitoring, employment agreements),
gpt-5.2's positive RDI is the safer profile: its errors are
visible additions that reviewers can identify and reject. In legal
operations workflows where false positives consume review capacity
and erode trust in the system, the ordering reverses. No single
model is universally correct; the appropriate choice depends on
the asymmetry of the legal task. RDI makes that choice tractable.

\section{Results: Calibrated Mitigation (Experiment 2)}
\label{sec:results-exp2}

\begin{figure*}[!t]
\centering
\includegraphics[width=0.88\textwidth]{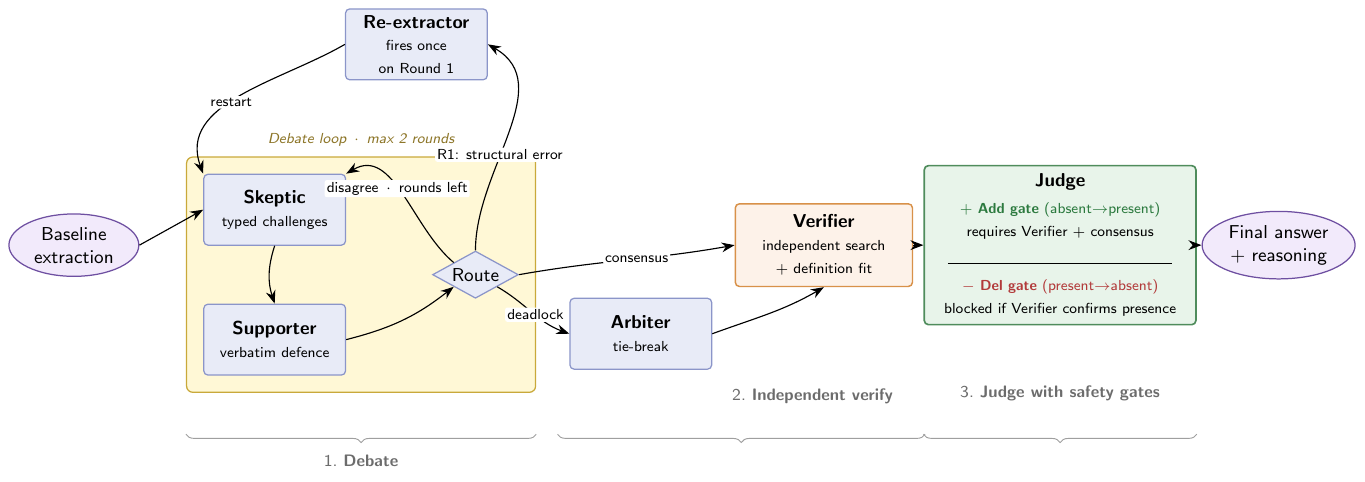}
\caption{Typed debate pipeline, organised into three phases.
\textbf{(1) Debate}: a Skeptic issues claim-type-specific
challenges (Appendix~\ref{app:skeptic}); a Supporter defends with
verbatim contract quotes; a Route node directs traffic. If the
Skeptic flags a structural error in Round~1, the Re-extractor
fires once and the loop restarts. If agents disagree with rounds
remaining, the loop continues; on deadlock, the Arbiter
tie-breaks conservatively.
\textbf{(2) Independent verify}: the Verifier searches the
contract independently and checks definition fit.
\textbf{(3) Judge with safety gates}: the Add gate
(absent~$\to$~present) requires both Verifier confirmation and
debate consensus, blocking fabricated additions; the Del gate
(present~$\to$~absent) is blocked when the Verifier confirms
presence, preventing erasure of correct findings. The asymmetry
encodes the measured FAR~$>$~FRR profile from Experiment~1.}
\label{fig:pipeline}
\end{figure*}

\begin{figure*}[!t]
\centering
\begin{minipage}[t]{0.46\textwidth}
\centering
\includegraphics[width=\linewidth]{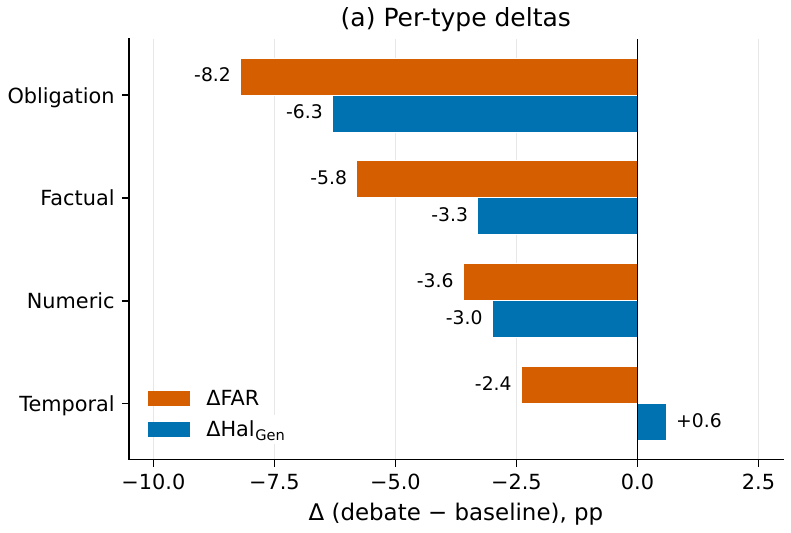}
\caption{Per-type deltas from Experiment~2. Gains concentrate on
obligation ($\Delta$FAR~$=-8.2$, $\Delta\mathrm{Hal_{Gen}}$~$=-6.3$)
and factual ($\Delta$FAR~$=-5.8$). Temporal $\mathrm{Hal_{Gen}}$ is
essentially unchanged ($+0.6$ pp), consistent with temporal being
the lowest-hallucination type at baseline. The calibrated
intervention produces the per-type pattern predicted by
Experiment~1. $\Delta$Hal~ in the legend denotes $\Delta\mathrm{Hal_{Gen}}$~}
\label{fig:deltas}
\end{minipage}
\hfill
\begin{minipage}[t]{0.46\textwidth}
\centering
\includegraphics[width=\linewidth]{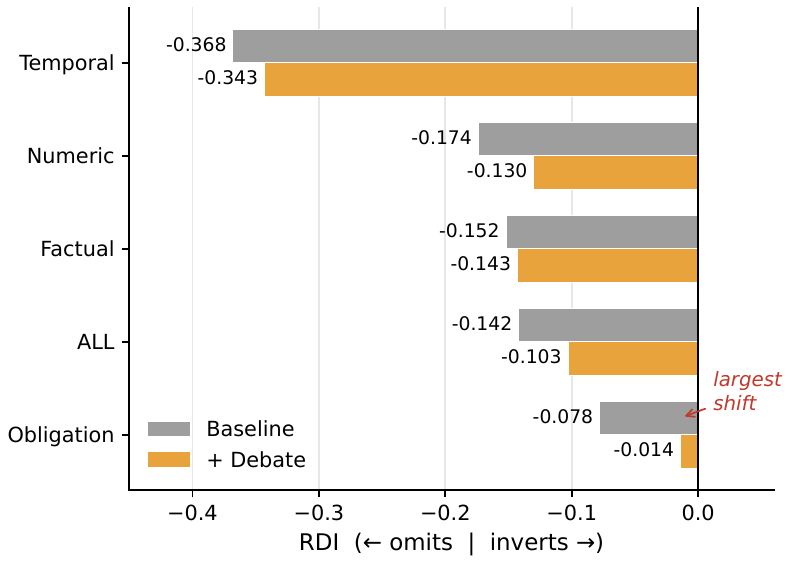}
\caption{RDI shift for gemma-4-26B-A4B after applying the typed debate
pipeline. The obligation category shows the largest correction
($-0.078 \to -0.014$, near-balanced). Skeptic challenges target
missing conditions and carve-outs, addressing the omission bias
Experiment~1 identified as the dominant obligation error direction.}
\label{fig:rdi_shift}
\end{minipage}
\end{figure*}

\subsection{From Typed Diagnosis to Calibrated Intervention}

Experiment~1 produces an actionable failure profile. Numeric and
obligation claims fail most. Parametric priors from pretraining
override explicit extraction instructions on threshold values.
Models that score equivalently on aggregate $\mathrm{Hal_{TP}}$ may nonetheless
fail in opposite compliance directions. This characterisation
doubles as a specification: a mitigation should reduce the FAR on
numeric and obligation clauses; it should compensate for
prior-substitution rather than ignoring it; and its effect on error
direction should be measurable.

The specific question for Experiment~2 is therefore narrow: can a
debate pipeline calibrated to the measured failure profile reduce
fabrication on a low-cost open model, and do the gains concentrate
on the highest-failure categories as the calibration predicts?

Figure~\ref{fig:pipeline} shows the pipeline; \S\ref{sec:method-debate}
describes its components in full. Skeptic challenges are calibrated
against the Experiment~1 per-type failure profile; asymmetric gates
encode the measured FAR~$>$~FRR risk asymmetry.

\subsection{Results: Fabrication Filtered, Direction Corrected}

\begin{table}[h]
\caption{Matched-subset comparison (120 contracts, run\_id$=$1).
$\mathrm{Hal_{Gen}}$ is the stricter generation-side metric
$(\mathrm{contradicted}+\mathrm{FP})/(\mathrm{TP}+\mathrm{FP})$,
penalising fabrications in addition to content errors. Score~$=$
mean rank across FAR, FRR, Acc, $\mathrm{Hal_{Gen}}$, JEq
(lower~$=$~better). qwen3-32b: 4{,}817 rows vs 4{,}920 nominal due
to export variation. Comparisons involving qwen3-32b on this subset should be interpreted with this row-count caveat.}
\label{tab:leaderboard}
\footnotesize\centering\setlength{\tabcolsep}{3pt}
\begin{tabular}{@{}clrrrrrr@{}}
\toprule
\# & Model & FAR & FRR & Acc & $\mathrm{Hal_{Gen}}$ & JEq & Sc. \\
\midrule
1 & gemma-debate   &  8.4 & 14.4 & 89.7 & 58.6 & 43.3 & \textbf{2.4} \\
2 & gpt-5.2        & 10.7 & 12.0 & 88.9 & 61.0 & 43.7 & 2.6 \\
3 & qwen3-32b      & 14.9 & 10.6 & 86.5 & 64.6 & 43.4 & 3.4 \\
4 & llama-3.3-70b  &  7.6 & 17.9 & 89.2 & 63.4 & 36.3 & 3.6 \\
5 & gemini-3-flash & 19.1 &  4.5 & 85.4 & 66.2 & 46.8 & 3.8 \\
6 & gemma-base & 15.4 & 15.8 & 84.5 & 64.8 & 41.8 & 5.2 \\
\bottomrule
\end{tabular}
\end{table}

The typed debate moves gemma-4-26B-A4B from last place (Score~5.2) to
first (Score~2.4) on the matched subset (rank~1 under 4 of 5
weighting schemes; gpt-5.2 leads under recall-heavy weighting,
App.~\ref{app:robustness}). The mechanism is fabrication filtering
rather than content correction: false-positive extractions drop
from 524 to 287 ($-45\%$) while content contradictions move only
642 to 641 ($-0.2\%$). The Skeptic can verify clause
\emph{existence} through absence-of-evidence reasoning, but is less effective at correcting \emph{content} errors within genuinely
present clauses because the baseline extraction and Skeptic share the
same parametric priors. This is consistent with~\cite{huang2024cannot}
and calibrates deployment expectations: typed debate reduces
fabrications but does not reliably repair what a present clause
says.

Figure~\ref{fig:deltas} validates the diagnostic predictions from Experiment 1. The typed
intervention predicts that obligation and factual claims will show
the largest gains (highest baseline FAR, with Skeptic challenges
most directly targeted to their failure modes) and that temporal
will show the smallest (lowest baseline FAR, and values that are
difficult to fabricate verbatim). The observed deltas match this
ordering: obligation $\Delta$FAR~$=-8.2$, factual $\Delta$FAR~$=-5.8$,
numeric $\Delta$FAR~$=-3.6$, and temporal $\Delta$FAR~$=-2.4$. The
ordering was specified in advance of running the mitigation rather
than read off the results, so the match provides evidence that the
typed diagnosis is informative beyond the summary it supplies.

Two facts in Table~\ref{tab:leaderboard} are decisive: gemma-debate
clears the commercial frontier on composite score (2.4 vs gpt-5.2
at 2.6), and the gap between gemma-base (5.2) and gemma-debate
(2.4) is the intervention's effect, holding the underlying model
fixed.

The corresponding direction correction appears in the obligation
RDI (Figure~\ref{fig:rdi_shift}): typed Skeptic challenges targeting
missing conditions, dropped carve-outs, and scope loss move
gemma-4-26B-A4B from omission-heavy ($-0.078$) to near-balanced
($-0.014$) on obligation claims. The challenge questions were
specified in advance to counteract omission bias because
Experiment~1 identified omission as the dominant obligation error
direction, so this shift is the intended consequence of the
calibration rather than an incidental effect.

\section{Discussion}

\paragraph{Deployment and governance implications.}
The $\sim$40~pp typed gap means any legal AI evaluation reporting
only aggregate $\mathrm{Hal_{TP}}$ averages a 29--35\% failure
rate on temporal claims alongside a 65--74\% rate on claims
determining liability thresholds and obligation scope. Two systems
scoring identically on $\mathrm{Hal_{TP}}$ can carry opposite risk
profiles,  a distinction RDI surfaces as a single comparable number.
For compliance workflows where missed obligations create liability,
a positive or near-zero RDI is the safer profile; for
legal-operations settings where false positives consume review
capacity, the ordering reverses. Typed profiles and RDI are
derivable from any oracle-bounded legal corpus, supporting typed
audits before deployment rather than relying on vendor-reported
aggregate accuracy.\

\textbf{Scope.}
The typed profiles and RDI values reported here apply to
CUAD-style English US commercial contracts. Whether the failure
ordering transfers to other document types is an empirical
question this paper does not resolve. What transfers is the
auditing method: any legal task with a verifiable oracle can
instantiate typed profiles and RDI, but resulting numbers will
differ. Practitioners should commission task-specific audits
rather than applying CUAD-derived thresholds to new contexts.

\section{Conclusion}

\textbf{LegalHalluLens} measures, on 249{,}252 clause-level instances
across four models, a consistent $\sim$40~pp hallucination gap (range 38.0--40.6~pp across models)
between obligation/numeric and temporal claims that aggregate
evaluation conceals. Two models with matched $\mathrm{Hal_{TP}}$
carry opposite risk profiles, operationalised by the Risk
Direction Index. A typed debate pipeline reduces fabricated
detections by 45\%, with per-category gains tracking the prior
diagnosis. The useful question for trustworthy legal AI deployment
is not the model's aggregate accuracy but which claim types it
fails on and, when it fails, in which direction.

\section*{Limitations}

Numerical results apply to 510 English-US commercial contracts
from CUAD; the typed failure ordering is consistent across four
architectures, but generalisation across jurisdictions and
document types remains to be verified. All experiments assume
full-document context; for contracts exceeding model context
windows, retrieval-augmented variants introduce additional failure
modes orthogonal to those measured here. Experiment~2 uses one run
with one backbone (gemma-4-26B-A4B) on a 120-contract subset; the
composite ranking is evidence for that comparison only.
Minimal-prompt and generic-debate ablations are direct extensions.

\textbf{Judge dependence.} All reported $\mathrm{Hal_{TP}}$,
$\mathrm{Hal_{Gen}}$, and RDI numbers flow through a single LLM
evaluation judge (gemini-2.5-flash) applying the rubric in
Appendix~\ref{app:judge}. The judge is held fixed and independent
of every extractor evaluated, and we have framed RDI as a
directional signal (\S\ref{sec:method-rdi}) precisely because the
absence of human-validated judge labels means small RDI
differences should not be over-interpreted; large bootstrap-stable
separations such as gpt-5.2 ($+0.161$) versus qwen3-32b
($-0.202$) lie well outside any plausible judge-noise band, but
per-category RDI values close to zero warrant additional caution.
Validating judge labels against expert annotation on a
stratified sample is a direct extension that would tighten the
cardinal interpretation of RDI without changing the directional
ordering.

\section*{Impact Statement}

\textbf{Diagnostic, not clearance.} Typed profiles provide finer
resolution than aggregate rates, supporting model comparison,
risk-aware deployment, and mitigation design. Even our best
configuration contradicted the source in 58.6\% of detected
clause contents, so typed evaluation should inform, not
replace, qualified human review in high-stakes legal workflows.

\textbf{Direction-aware deployment.} The RDI surfaces a systematic
bias that aggregate metrics conceal. Compliance workflows (where
missed obligations create liability) benefit from systems with a
positive or near-zero RDI; legal-operations settings (where false
positives consume review capacity) may prefer the opposite
profile. The framework makes this trade-off legible.

\textbf{Agent design and dual-use.} Calibrated multi-agent
extraction pipelines could be misused to produce the
\emph{appearance} of compliance review without the substance,
e.g., automated due-diligence reports that meet a procedural bar
while masking the 40+~pp typed gap. We recommend against
autonomous deployment without (i) per-deployment re-measurement
of the typed profile on representative documents, (ii)
human-in-the-loop review of all flagged clauses in obligation and
numeric categories, and (iii) explicit disclosure that aggregate
accuracy does not bound legal risk. The diagnostic framework
itself is intended to support, not replace, this oversight.

\textbf{Scope of evidence.} Numerical results apply to CUAD-style
English US commercial contracts. The methodology extends to any
legal task with a verifiable source, but specific failure rates
should be re-measured for each new deployment context.
\textbf{LLM usage.} The authors used Claude Opus~4.6 and Claude
Sonnet~4.6 (Anthropic) for writing assistance (drafting, polishing,
grammar, literature reading) and code assistance (scaffolding,
debugging). Research design, methodology, and conclusions are the
authors' own work; the authors take full responsibility for all
content.

\section*{Code Availability}
Code, prompts, and analysis scripts are available at   \url{https://github.com/lalitdv9/LegalHallulens}.

\bibliography{LegalHalluLens}

@inproceedings{bang2025hallulens,
  author    = {Bang, Yejin and Ji, Ziwei and Schelten, Alan and Hartshorn, Anthony and Fowler, Tara and Zhang, Cheng and Cancedda, Nicola and Fung, Pascale},
  title     = {{HalluLens}: {LLM} Hallucination Benchmark},
  booktitle = {Proceedings of the 63rd Annual Meeting of the Association for Computational Linguistics},
  year      = {2025}
}

@inproceedings{ji2025phantom,
  author    = {Ji, Lanlan and Seyler, Dominic and Kaur, Gunkirat and Hegde, Manjunath and Dasgupta, Koustuv and Xiang, Bing},
  title     = {{PHANTOM}: A Benchmark for Hallucination Detection in Financial
               Long-Context {QA}},
  booktitle = {Proceedings of NeurIPS Datasets and Benchmarks},
  year      = {2025}
}

@inproceedings{min2023factscore,
 author = {Min, Sewon and Krishna, Kalpesh and Lyu, Xinxi and Lewis, Mike and Yih, Wen-tau and Koh, Pang Wei and Iyyer, Mohit and Zettlemoyer, Luke and Hajishirzi, Hannaneh},
  title     = {{FActScore}: Fine-Grained Atomic Evaluation of Factual Precision
               in Long Form Text Generation},
  booktitle = {Proceedings of EMNLP},
  year      = {2023}
}

@article{ravi2024halubench,
  author    = {Ravi, Selvan Sunitha and Mielczarek, Bartosz and Kannappan, Anand and Kiela, Douwe and Qian, Rebecca},
  title     = {Lynx: An Open Source Hallucination Evaluation Model},
  journal   = {arXiv preprint arXiv:2407.08488},
  year      = {2024}
}

@article{guha2023legalbench,
  author = {Guha, Neel and Nyarko, Julian and Ho, Daniel E. and R{\'e}, Christopher and Chilton, Adam and Narayana, Aditya and Chohlas-Wood, Alex and Peters, Austin and Waldon, Brandon and Rockmore, Daniel N. and others},
  title     = {{LegalBench}: A Collaboratively Built Benchmark for Measuring
               Legal Reasoning in Large Language Models},
  journal   = {arXiv preprint arXiv:2308.11462},
  year      = {2023}
}

@inproceedings{blairstanek2024blt,
  author    = {Blair-Stanek, Andrew and Holzenberger, Nils and Van Durme, Benjamin},
  title     = {{BLT}: Can Large Language Models Handle Basic Legal Text?},
  booktitle = {Proceedings of the Natural Legal Language Processing Workshop},
  pages     = {216--232},
  year      = {2024}
}

@inproceedings{liu2025contracteval,
  author    = {Liu, Shuang and Li, Zelong and Ma, Ruoyun and Zhao, Haiyan and Du, Mengnan},
  title     = {{ContractEval}: Benchmarking {LLMs} for Clause-Level Legal Risk
               Identification in Commercial Contracts},
  booktitle = {Proceedings of the Natural Legal Language Processing Workshop},
  year      = {2025},
  doi       = {10.18653/v1/2025.nllp-1.19}
}

@inproceedings{hou2024gaps,
  author    = {Hou, Abe Bohan and Jurayj, William and Holzenberger, Nils and Blair-Stanek, Andrew and Van Durme, Benjamin},
  title     = {Gaps or Hallucinations? {S}crutinizing Machine-Generated Legal
               Analysis for Fine-Grained Text Evaluations},
  booktitle = {Proceedings of the Natural Legal Language Processing Workshop},
  year      = {2024}
}

@article{dahl2024fictions,
  author    = {Dahl, Matthew and Magesh, Varun and Suzgun, Mirac and Ho, Daniel E.},
  title     = {Large Legal Fictions: {P}rofiling Legal Hallucinations in Large
               Language Models},
  journal   = {Journal of Legal Analysis},
  volume    = {16},
  number    = {1},
  pages     = {64--93},
  year      = {2024}
}

@article{magesh2024hallucinationfree,
  author    = {Magesh, Varun and Surani, Faiz and Dahl, Matthew and Suzgun, Mirac
               and Manning, Christopher D. and Ho, Daniel E.},
  title     = {Hallucination-Free? {A}ssessing the Reliability of Leading {AI}
               Legal Research Tools},
  journal   = {Journal of Empirical Legal Studies},
  volume    = {22},
  pages     = {216--242},
  year      = {2025},
  doi       = {10.1111/jels.12413}
}

@inproceedings{demir2025validate,
  author    = {Demir, M. Mikail and Canbaz, M. Abdullah},
  title     = {Validate Your Authority: {B}enchmarking {LLMs} on Multi-Label
               Precedent Treatment Classification},
  booktitle = {Proceedings of the Natural Legal Language Processing Workshop},
  year      = {2025}
}

@inproceedings{enguehard2025lemaj,
  author    = {Enguehard, Joseph and Van Ermengem, Morgane and Atkinson, Kate and Cha, Sujeong and Ghosh Chowdhury, Arijit and Kallur Ramaswamy, Prashanth and Roghair, Jeremy and Marlowe, Hannah R. and Negreanu, Carina Suzana and Boxall, Kitty and Mincu, Diana},
  title     = {{LeMAJ} (Legal {LLM}-as-a-Judge): {B}ridging Legal Reasoning
               and {LLM} Evaluation},
  booktitle = {Proceedings of the Natural Legal Language Processing Workshop},
  year      = {2025},
  doi       = {10.18653/v1/2025.nllp-1.23}
}

@inproceedings{hendrycks2021cuad,
author = {Hendrycks, Dan and Burns, Collin and Chen, Anya and Ball, Spencer},
title = {{CUAD}: An Expert-Annotated {NLP} Dataset for Legal Contract Review},
  booktitle = {Proceedings of NeurIPS},
  year      = {2021}
}

@inproceedings{du2024multiagent,
  author    = {Du, Yilun and Li, Shuang and Torralba, Antonio and Tenenbaum, Joshua B. and Mordatch, Igor},
  title     = {Improving Factuality and Reasoning in Language Models through
               Multiagent Debate},
  booktitle = {Proceedings of ICML},
  pages     = {11733--11763},
  year      = {2024}
}

@inproceedings{fang2025counterfactual,
  author    = {Fang, Yi and Li, Moxin and Wang, Wenjie and Hui, Lin and Feng, Fuli},
  title     = {Counterfactual Debating with Preset Stances for Hallucination
               Elimination of {LLMs}},
  booktitle = {Proceedings of COLING},
  pages     = {10554--10568},
  year      = {2025}
}

@inproceedings{liu2025selfdebating,
  author    = {Li, Miaoran and Chen, Jiangning and Xu, Minghua and Wang, Xiaolong},
  title     = {Hallucination Detection in Structured Query Generation via
               {LLM} Self-Debating},
  booktitle = {Findings of EMNLP},
  pages     = {16102--16113},
  year      = {2025}
}

@inproceedings{hu2025debategraph,
  author    = {Hu, Wentao and Zhang, Wengyu and Jiang, Yiyang and Zhang, Chen Jason and Wei, Xiaoyong and Li, Qing},
  title     = {Removal of Hallucination on Hallucination: {D}ebate-Augmented
               {RAG}},
  booktitle = {Proceedings of ACL},
  pages     = {15839--15853},
  year      = {2025}
}

@article{snell2024scaling,
  author    = {Snell, Charlie and Lee, Jaehoon and Xu, Kelvin and Kumar, Aviral},
  title     = {Scaling {LLM} Test-Time Compute Optimally Can Be More Effective
               than Scaling Model Parameters},
  journal   = {arXiv preprint arXiv:2408.03314},
  year      = {2024}
}

@article{wu2024inference,
  author    = {Wu, Yangzhen and Sun, Zhiqing and Li, Shanda and Welleck, Sean and Yang, Yiming},
  title     = {Inference Scaling Laws: An Empirical Analysis of Compute-Optimal
               Inference for Problem-Solving with Language Models},
  journal   = {arXiv preprint arXiv:2408.00724},
  year      = {2024}
}

@inproceedings{huang2024cannot,
  author    = {Huang, Jie and Chen, Xinyun and Mishra, Swaroop and Zheng, Huaixiu Steven and Yu, Adams Wei and Song, Xinying and Zhou, Denny},
  title     = {Large Language Models Cannot Self-Correct Reasoning Yet},
  booktitle = {Proceedings of ICLR},
  year      = {2024}
}

@inproceedings{purushothama2025bench,
  author    = {Purushothama, Abhishek and Min, Junghyun and Waldon, Brandon and Schneider, Nathan},
  title     = {Not Ready for the Bench: {LLM} Legal Interpretation Is Unstable
               and Uncalibrated to Human Judgments},
  booktitle = {Proceedings of the Natural Legal Language Processing Workshop},
  year      = {2025},
  doi       = {10.18653/v1/2025.nllp-1.22}
}
\bibliographystyle{icml2026}

\clearpage\appendix\onecolumn

\section{Judge Prompt}
\label{app:judge}
The following is the verbatim prompt used by the external evaluation
judge (gemini-2.5-flash, temperature~$=$~0). It is applied
identically across all four extraction backbones in Experiment~1
and across all six configurations in Experiment~2.
\begin{verbatim}
You are an expert legal contract verifier.

TASK:
Determine whether ANSWER 1 (AI) is semantically equivalent to
ANSWER 2 (Ground Truth) for the SAME clause and contract.

CLAUSE: {clause_name}

ANSWER 1 (AI Generated):
{ai_answer}

ANSWER 2 (Ground Truth):
{gt_answer}

DECISION CRITERIA (BE STRICT ON PRECISION):
Return "equivalent": true ONLY IF all of the following hold:
1) CORE FACTS MATCH: The same parties/actors,
   rights/obligations, and conditions are stated.
2) NUMERIC PRECISION MATCHES: Any amounts, percentages,
   thresholds, caps, quantities, and units (including time basis
   like per month/per year) are the same. Any mismatch
   => equivalent=false.
3) TEMPORAL PRECISION MATCHES: Any dates, durations, notice
   periods, renewal terms, survival periods, and timelines are
   the same. Any mismatch => equivalent=false.
4) MODALITY/POLARITY MATCHES: must/shall vs may, prohibited vs
   permitted, and any negation (not/unless/except) must match.
   Any mismatch => equivalent=false.
5) EXCEPTIONS/CARVE-OUTS: If either answer includes an exception,
   carve-out, or condition, the other must include the same
   exception/condition in substance.
   Otherwise => equivalent=false.

ALLOWABLE DIFFERENCES:
- Formatting, whitespace, and punctuation.
- Reordering of equivalent statements.
- Minor paraphrases that do not change any of the precise facts
  above.

OUTPUT (JSON ONLY):
Return ONLY a valid JSON object:
{
  "equivalent": true/false,
  "reason": "one short sentence",
  "mismatch_type": "none|numeric|temporal|obligation|scope|
                    missing_condition|extra_condition|other"
}

RULES:
- If either answer is empty or says "Not present" while the
  other contains content, equivalent=false.
- If the AI answer is a subset of the ground truth but misses a
  required condition/exception, equivalent=false.
- Do not add any extra text outside the JSON.
\end{verbatim}
\texttt{missing\_condition}: AI omits a carve-out or condition
present in ground truth.
\texttt{extra\_condition}: AI asserts an obligation, condition,
or qualifier absent from the source.

\section{Extraction Prompt (abbreviated)}
\label{app:extract}
\begin{verbatim}
You are a legal AI assistant analyzing a commercial contract.
Use ONLY the provided contract text. No outside knowledge.

For EACH of the 41 CUAD clause types:
- If present: return ALL spans capturing the operative meaning,
  including exceptions, carve-outs, conditions, notice periods,
  and cross-references ("subject to","except","provided that").
- If not present: mark is_impossible=true, answer=[].

SELF-CHECK: Re-scan for additional conditions, numeric thresholds,
and cross-references before finalising output.

OUTPUT: JSON array of 41 items:
{ "clause_name": str, "is_impossible": bool, "answer": [str] }
Complete ALL 41. Temperature = 0.
\end{verbatim}

\subsection*{B.1 Numeric Clause Definitions (verbatim)}
\label{app:numeric-defs}
The five numeric clause types are defined to the model with the
following NOTE blocks specifying exclusions. These are referenced
in \S\ref{sec:results-exp1} as evidence that pretraining priors
override explicit prompt guidance.

\begin{verbatim}
"Cap On Liability": Does the contract include a cap on
liability upon the breach of a party's obligation? This
includes time limitation for the counterparty to bring claims
or maximum amount for recovery. NOTE: This requires an
explicit maximum amount or formula capping liability. A clause
that ONLY excludes certain types of damages (e.g. no
consequential damages) without stating a maximum liability
amount is typically not a Cap On Liability.

"Liquidated Damages": Does the contract contain a clause that
would award either party liquidated damages for breach or a
fee upon the termination of a contract? NOTE: The clause must
AWARD or SPECIFY a liquidated damages amount. A clause
EXCLUDING or DENYING liability for liquidated damages (e.g.
"no liability for liquidated damages") is the OPPOSITE - it is
NOT a Liquidated Damages clause.

"Minimum Commitment": Is there a minimum order size or minimum
amount or units per-time period that one party must buy from
the counterparty? NOTE: This includes purchase minimums, order
minimums, AND performance minimums. A recurring fixed service
fee where no minimum quantity is specified is less likely to
qualify.

"Volume Restriction": Is there a fee increase or consent
requirement if one party's use exceeds certain threshold?
NOTE: This is an explicit MAXIMUM CAP or threshold on
usage/quantity that triggers a fee or consent requirement.
Minimum purchase quotas are NOT Volume Restrictions.

"Price Restrictions": Is there a restriction on the ability
of a party to raise or reduce prices? NOTE: This restricts the
PRICING DISCRETION of a party - their ability to SET or CHANGE
prices. A payment cap or maximum payment amount is a PAYMENT
LIMIT, not a Price Restriction.
\end{verbatim}
\noindent The remaining 36 clause definitions follow the same
pattern.

\section{Typed Skeptic Challenge Questions}
\label{app:skeptic}

Challenge questions are derived from the dominant failure mode per
type identified in Experiment~1, not from general-purpose
verification heuristics.

\textbf{Numeric (5 types).} Is this exact value stated verbatim
in the contract, or is it a plausible prior assumption about
common threshold values for this clause type? Is the unit of
measurement explicit and correct (per month vs per year; USD vs
percentage)? Is any cap, floor, or qualifier (``up to'', ``at
least'', ``not to exceed'') present in the contract but absent
from the extraction?

\textbf{Obligation/Entitlement (27 types).} What is the exact
modal verb in the contract (shall/must/may/should/will), does
the extraction preserve it, or has it been upgraded or downgraded?
Are ALL trigger conditions and antecedents that must occur before
this obligation activates captured? Are there exceptions,
carve-outs, or ``provided that / except / unless'' clauses in the
text that the extraction omits? Is any geographic, temporal, or
subject-matter scope limitation dropped?

\textbf{Temporal (6 types).} Is the date or duration stated
explicitly and verbatim, or inferred from surrounding context?
Is the notice period unit exact (30 days is not equivalent to one
month)? Could this be a common boilerplate value assumed from
prior rather than read from this specific contract?

\textbf{Factual (3 types).} Is this fact explicitly stated in
the contract text, or is the model drawing on outside knowledge?
Is the exact legal entity name used as it appears in the contract?

\section{CUAD Clause-to-Category Mapping}
\label{app:mapping}

\textbf{Numeric (5):} Cap on Liability; Minimum Commitment; Volume
Restriction; Price Restrictions; Liquidated Damages.

\textbf{Temporal (6):} Agreement Date; Effective Date; Expiration
Date; Renewal Term; Notice Period to Terminate Renewal; Warranty
Duration.

\textbf{Obligation/Entitlement (27):} Non-Compete; Exclusivity;
No-Solicit of Customers; No-Solicit of Employees; License Grant;
IP Ownership Assignment; Joint IP Ownership; Non-Transferable
License; Audit Rights; Insurance; Termination for Convenience;
Post-Termination Services; Most Favored Nation; Competitive
Restriction Exception; Non-Disparagement; Rofr/Rofo/Rofn; Change
of Control; Anti-Assignment; Revenue/Profit Sharing; Affiliate
License-Licensor; Affiliate License-Licensee;
Unlimited/All-You-Can-Eat-License; Irrevocable or Perpetual
License; Source Code Escrow; Uncapped Liability; Covenant Not to
Sue; Third Party Beneficiary.

\textbf{Factual (3):} Document Name; Parties; Governing Law.

\section{Robustness Analyses}
\label{app:robustness}

This appendix reports robustness checks supporting claims in the
main text. All analyses use the same data as Experiments~1 and~2.

\subsection*{E.1 Per-Run Variance (Experiment 1)}

Standard deviations across the three independent runs are small
relative to the within-model typed gap (38.0--40.6~pp), confirming
that the typed ordering is a stable property rather than run noise.

\begin{center}\footnotesize
\begin{tabular}{@{}lccccc@{}}
\toprule
Model & FAR & FRR & Acc & $\mathrm{Hal_{TP}}$ & JEq \\
\midrule
gemini-3-flash & 19.1 (0.1) & 4.5 (0.0) & 85.6 (0.0) & 50.9 (0.0) & 46.9 (0.0) \\
gpt-5.2        & 11.8 (0.2) & 11.6 (0.1) & 88.3 (0.1) & 51.9 (0.4) & 42.6 (0.3) \\
qwen3-32b      & 13.4 (0.4) & 10.8 (0.3) & 87.5 (0.2) & 52.1 (0.6) & 42.7 (0.7) \\
llama-3.3-70b  & 7.7 (0.1)  & 18.0 (0.1) & 89.0 (0.1) & 56.5 (0.2) & 35.7 (0.2) \\
\bottomrule
\end{tabular}
\end{center}
\noindent Mean (SD) across 3 runs. Largest SD on $\mathrm{Hal_{TP}}$ is 0.6~pp.
Per-category $\mathrm{Hal_{TP}}$ SDs are $\leq 2.4$~pp (largest:
gpt-5.2 numeric). All within-model typed gaps remain
$\geq 36$~pp at the 1-SD bound.

\subsection*{E.2 RDI Bootstrap CIs by Category}

95\% bootstrap confidence intervals (2{,}000 resamples) over all
runs pooled. Aggregate (ALL) intervals appear in §\ref{sec:direction};
the typed breakdown shows the directional separation holds within
the dominant Obligation category, where the ordering is
deployment-relevant.

\begin{center}\footnotesize
\begin{tabular}{@{}lcc@{}}
\toprule
Model & RDI [95\% CI] & Category \\
\midrule
gpt-5.2 & $+0.220$ \footnotesize{$[+0.207, +0.234]$} & obligation \\
gemini-3-flash & $+0.018$ \footnotesize{$[+0.004, +0.031]$} & obligation \\
llama-3.3-70b & $-0.198$ \footnotesize{$[-0.210, -0.186]$} & obligation \\
qwen3-32b & $-0.181$ \footnotesize{$[-0.194, -0.168]$} & obligation \\
\bottomrule
\end{tabular}
\end{center}

\subsection*{E.3 Composite Rank Sensitivity}

Composite Score (\S\ref{sec:results-exp2}) uses equal weights across FAR,
FRR, Acc, $\mathrm{Hal_{Gen}}$, JEq. Robustness across alternative
weightings:

\begin{center}\footnotesize
\begin{tabular}{@{}lccc@{}}
\toprule
Scheme & gemma-debate & gpt-5.2 & gemma-base \\
\midrule
Equal (1:1:1:1:1)             & \textbf{1} & 2 & 6 \\
FP-heavy (2:1:1:2:1)          & \textbf{1} & 2 & 6 \\
Recall-heavy (1:2:1:1:2)      & 2          & \textbf{1} & 6 \\
Halluc-only (0:0:0:1:0)       & \textbf{1} & 2 & 5 \\
Detection-only (1:1:1:0:0)    & \textbf{1} & 2 & 6 \\
\bottomrule
\end{tabular}
\end{center}
\noindent gemma-debate ranks first under 4 of 5 schemes; gpt-5.2
leads under recall-heavy weighting. The intervention's improvement
over gemma-base (rank 5--6 in every scheme) is robust to
weighting.

\subsection*{E.4 Missing-Row Attribution}

Per-run row counts vs. the nominal 20{,}910 per run:

\begin{center}\footnotesize
\begin{tabular}{@{}lccc@{}}
\toprule
Model & Run 1 & Run 2 & Run 3 \\
\midrule
gemini-3-flash & 20860 & 20860 & 20860 \\
gpt-5.2        & 20910 & 20910 & 20869 \\
qwen3-32b      & 20417 & 20771 & 20348 \\
llama-3.3-70b  & 20843 & 20802 & 20802 \\
\bottomrule
\end{tabular}
\end{center}
\noindent For qwen3-32b (the largest variation): 5 contracts are
incomplete in all 3 runs; 58 contracts are incomplete in any run;
persistent fraction 8.6\%. Variation is contract-correlated, not
random,  a small set of inputs the model consistently fails to
process under temperature~$=$~0 API calls.

\subsection*{E.5 Obligation Subtype Profiles}

The Obligation/Entitlement category aggregates 27 CUAD types. Mean
$\mathrm{Hal_{TP}}$ across all four models, by subtype:

\begin{center}\footnotesize
\begin{tabular}{@{}lr@{\hskip 1.5em}lr@{}}
\toprule
Subtype & $\mathrm{Hal_{TP}}$ & Subtype & $\mathrm{Hal_{TP}}$ \\
\midrule
Post-Termination Services & 88.6 & Change of Control & 68.3 \\
Affiliate License-Licensor & 85.1 & Uncapped Liability & 66.1 \\
Source Code Escrow & 84.1 & Joint IP Ownership & 63.6 \\
IP Ownership Assignment & 80.5 & Non-Compete & 62.2 \\
Revenue/Profit Sharing & 79.3 & Covenant Not to Sue & 61.1 \\
Exclusivity & 79.1 & Irrev./Perpetual License & 60.5 \\
Affiliate License-Licensee & 79.0 & No-Solicit of Customers & 57.4 \\
Audit Rights & 77.2 & Anti-Assignment & 56.5 \\
Competitive Rest. Except. & 76.9 & No-Solicit of Employees & 54.1 \\
Non-Transferable License & 74.2 & Most Favored Nation & 48.3 \\
Unlimited/All-You-Can-Eat & 73.0 & Third Party Beneficiary & 42.9 \\
License Grant & 71.6 & Termination for Conv. & 42.7 \\
Non-Disparagement & 69.1 & & \\
Rofr/Rofo/Rofn & 69.0 & & \\
Insurance & 68.8 & & \\
\bottomrule
\end{tabular}
\end{center}
\noindent Range 42.7--88.6\%; within-bucket SD 12.4~pp. Even the
lowest obligation subtype (42.7\%) lies above the temporal
category mean (29.0--35.1\%), confirming the typed gap survives
intra-bucket heterogeneity.

\subsection*{E.6 Debate Pipeline Overhead}

Cost proxy for the typed debate pipeline (Experiment~2,
gemma-4-26B-A4B, 4{,}920 clause-level decisions):

\begin{center}\footnotesize
\begin{tabular}{@{}lcc@{}}
\toprule
Statistic & Value & Note \\
\midrule
Mean rounds                & 1.12 & max~$=$~2 \\
Median rounds              & 1    & 87.9\% finish in R1 \\
\% detections changed      & 12.8 & flips at gates \\
Skeptic--Supporter consensus & 99.94\% & \\
Stability (1.0)            & 90.6\% & remaining 9.4\% partial \\
\bottomrule
\end{tabular}
\end{center}
\noindent Per-type flip rates: factual 4.2\%, temporal 9.9\%,
numeric 13.8\%, obligation 14.2\% ,  consistent with the
per-category $\Delta$FAR ordering in Figure~\ref{fig:deltas}. Mean
rounds-per-type span only 1.10--1.20, indicating that the
calibrated benefit comes from \emph{which} clauses are flipped
rather than from extended deliberation.

\end{document}